\documentclass{article} 
\usepackage{iclr2015,times}
\usepackage{hyperref}
\usepackage{url}
\usepackage{graphicx}
\usepackage{epstopdf}
\usepackage{adjustbox}
\usepackage{xcolor}
\usepackage{array}

\title{Attention for Fine-Grained Categorization}

\author{
Pierre Sermanet, Andrea Frome, Esteban Real \\
Google, Inc.\\
\texttt{\{sermanet,afrome,ereal,\}@google.com}
}

\iclrfinalcopy 

\begin{document}

\maketitle

\begin{abstract}

This paper presents experiments extending the work
of ~\citet{BaICLR2015} on recurrent neural models for attention
into less constrained visual environments, specifically
fine-grained categorization on the Stanford Dogs data set.
In this work we use an RNN of the same structure but substitute
a more powerful visual network and perform large-scale
pre-training of the visual network outside of the attention
RNN. Most work in attention models to date focuses on tasks with toy or more
constrained visual environments, whereas we present
results for fine-grained categorization better than the state-of-the-art
GoogLeNet classification model. We show that our model learns to direct high
resolution attention to the most discriminative regions without any spatial
supervision such as bounding boxes, and it is able to discriminate
fine-grained dog breeds moderately well even when given only an initial low-resolution
context image and narrow, inexpensive glimpses at faces and fur patterns.
This and similar attention models have the major advantage of being trained
end-to-end, as opposed to other current detection and recognition pipelines
with hand-engineered components where information is lost. While our
model is state-of-the-art, further work is needed to fully
leverage the sequential input.

\end{abstract}


\section{Introduction}
\label{sec:intro}

This work presents experiments extending the work of~\citet{BaICLR2015}
on recurrent neural models for attention into less constrained visual
environments, specifically fine-grained
categorization on the Stanford Dogs data set. \citet{BaICLR2015} tackles the challenging problem of sequence
prediction in simplified visual settings (MNIST and Street View House Numbers)
using a recurrent model of attention similar to \citet{MnihNIPS2014}.
Complementary to that work, we are addressing the simpler task of classification
but in a visual environment with significant clutter and occlusion, variations
in lighting and pose, and a more difficult class discrimination task.
Previous work in learned visual
attention models has tackled a number of computer vision problems and
demonstrated the benefits of various attention mechanisms, though most of the
work has been focused on toy or more constrained environments,
such as detecting simple shapes \citep{Schmidhuber91},
tasks based on MNIST digits \citep{LarochelleNIPS2010, BazzaniICML2011,
DenilNeuralComp2012, RanzatoArxiv2014, MnihNIPS2014},
the vision-control game of ``catch'' \citep{MnihNIPS2014},
expression classification for 100$\times$100 aligned faces
\citep{LarochelleNIPS2010, ZhengIJCV2014},
detection of frontal faces \citep{TangArxiv2013},
tracking of hockey players \citep{BazzaniICML2011, DenilNeuralComp2012}
and gesture recognition \citep{DarrellNIPS1996}.
Most recently, two papers have explored attention mechanisms for more
complex visual input: \citet{GonzalezVF14} presented an attention-like model
for detection of chairs, people, cars, doors, and tables from SUN2012 images
\citep{SUN2012}; \citet{XuCaptioning2015} applied two different neural network
visual attention
models to the task of caption generation for MSCOCO images \citep{MSCOCO}.

Recent work in object detection \citep{SzegedyREA14, girshick2014rcnn} and
fine-grained categorization \citep{ZhangDGD14} use candidate object or part
proposals as a precursor to a more expensive classification stage. In these
systems, the proposals are generated from a bottom-up segmentation as
in \citet{girshick2014rcnn} or from a separate neural network as in
\citet{SzegedyREA14}.
As in our work, these pipelines are able to process
candidate regions for classification at a higher resolution and save processing
by focusing on a restricted set of regions.
Drawbacks to these systems are that they consist of independent processing steps
with engineered connections between them where information is lost. As an
example, they either do not aggregate evidence from independent candidate image
regions or do so with an ad hoc technique. They also lose
rich information between the candidate proposal and classification parts of the
pipeline. In contrast,
our models and others with similar structure are trained end-to-end to
incorporate information across observations, which can be expected to yield a
better final result.

\begin{figure}[h]
\begin{center}
\includegraphics[width=5in]{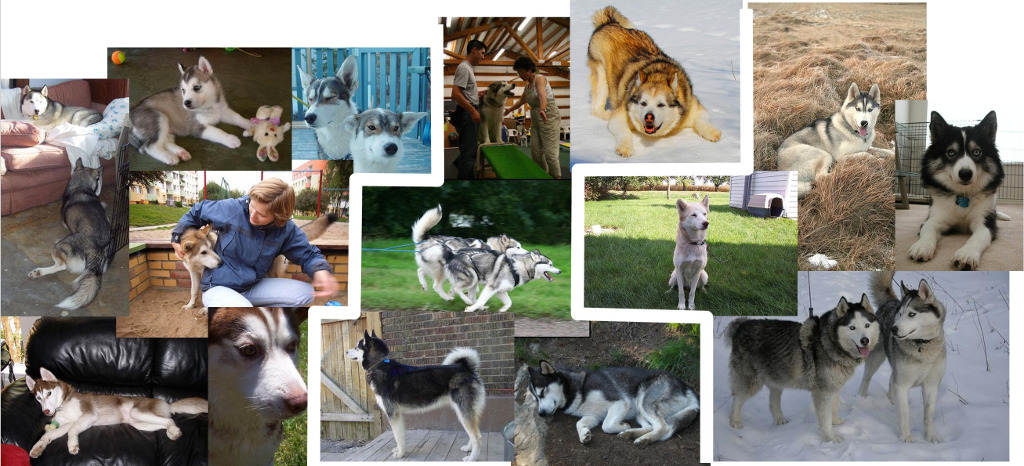}
\end{center}
\caption{Three classes from the Dogs data set that are difficult to tell
apart due to high intra-class variability and high similarity across
classes. The lines show the class boundaries; the classes are
Eskimo Dog on the left, Malamute in the center, and Siberian Husky on the
right. Of the 120 classes in Stanford Dogs, our model performs worst on
Siberian Husky.}
\label{fig:dogsAreHard}
\end{figure}

We apply the visual attention model from \citet{BaICLR2015} to the
Stanford Dogs fine-grained categorization task \citep{Dogs2011},
choosing
to perform the task without using the provided bounding boxes for
training or testing. This amounts to learning to simultaneously localize
and classify objects within scenes despite difficult class boundaries,
large variations in pose and lighting, varying and cluttered backgrounds, and
occlusion (Figure \ref{fig:dogsAreHard}). Fine-grained categorization is a
natural proving ground for attention-based
models. When performing classification at the sub-category level, e.g.
German Shepherd versus Poodle, the background is often
uncorrelated with class and acts as a distraction to the primary task.
As a result, several hand-crafted vision pipelines use provided bounding
boxes to isolate
the object of interest or may perform segmentation of the object
from the background, e.g. \citet{ParkhiCVPR2011, ChaiICCV2013,
AngelovaCVPR2013}.
Attention models could address this challenge
by learning to focus processing and discriminatory power on the parts of the
image that are relevant for the task without requiring expensive hand-labeled
bounding boxes.
In addition to ignoring the
distractors in the image, a good attention model could learn to focus
processing power on the specific features of the objects that help to
tell them apart, for example the face, ears, and particular fur patterns for dogs.
Future versions of this model could potentially also choose the scale at which
to examine details.


\section{Model Description}
\label{sec:model}

The structure of our model is nearly the same as that
presented in \citet{BaICLR2015} with a few differences; we give
an overview of their model here and describe the ways in which our model differs.
We refer the reader to that work for a more in-depth description of the network
choices and training procedure.

\begin{figure}
\begin{center}
\includegraphics[width=5in]{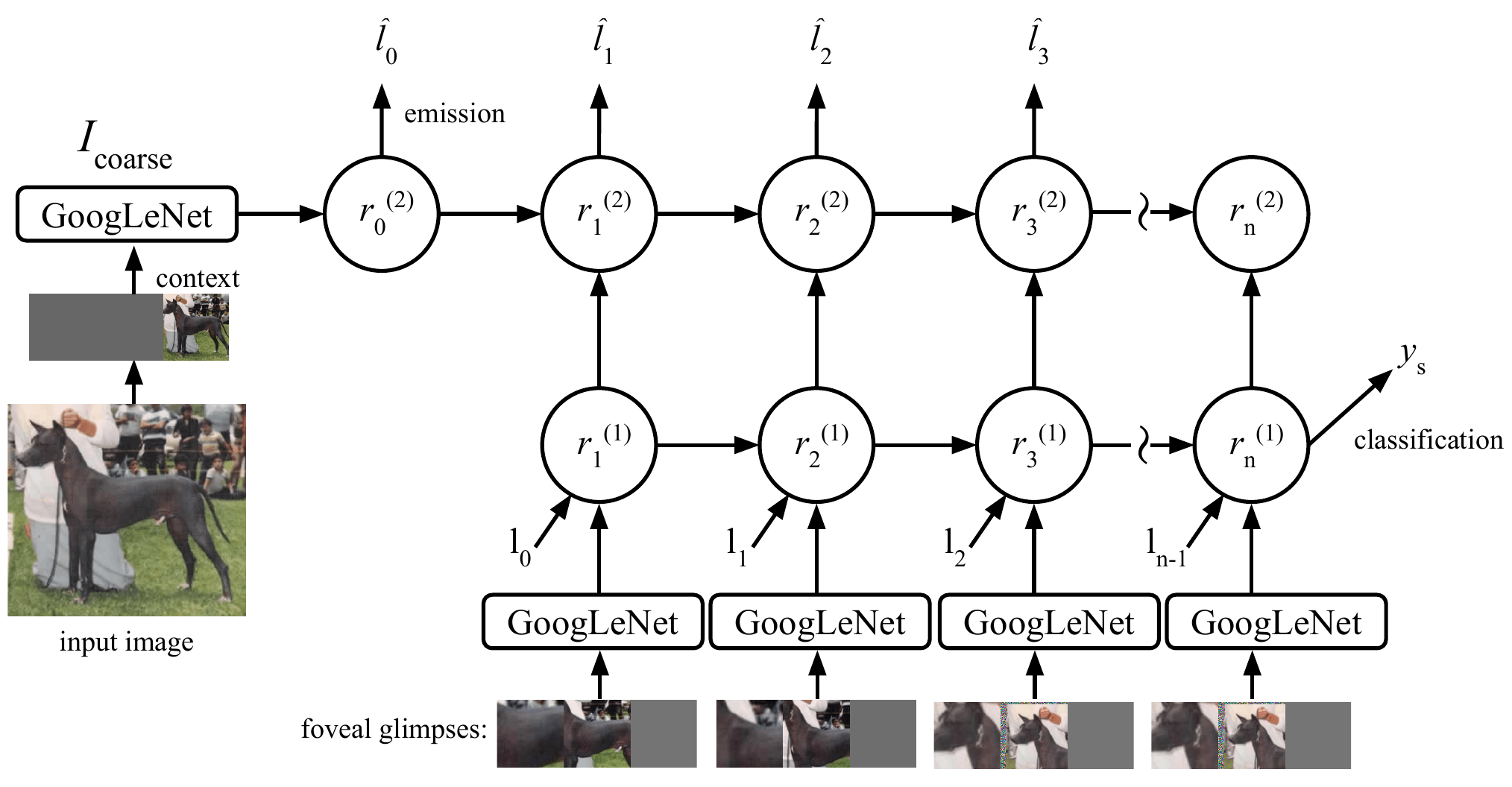}
\end{center}
\caption{Diagram of the model. The grayed-out boxes denote resolutions not in
use; in our experiments the context is always a low-resolution patch, while
each glimpse can be any combination of the low-, medium-, and high-resolution
patches.}
\label{fig:model}
\end{figure}

Figure \ref{fig:model} shows the structure of the model.
The system as a whole takes as input an image of any size and outputs N-way
classification scores using a softmax classifier, which is a similar
task to the finding digits and digit addition tasks in
\citet{BaICLR2015}.
The model is a recurrent neural network, with $N$ steps that correlate
with $N$ ``glimpses'' into the input image. At step $n$, the model receives
row and column coordinates $l_{n-1}$, which describe a point in the input image.
The network extracts a multi-resolution patch from the input
image at those coordinates, passes the pixels through fully-connected layers
which combine with activations from the previous glimpse step, and either
outputs coordinates $\hat{l}_n$ for the next glimpse or a final
classification $y_s$.

\begin{figure}
\begin{center}
\begin{tabular}{VVVV}
\includegraphics[scale=0.25]{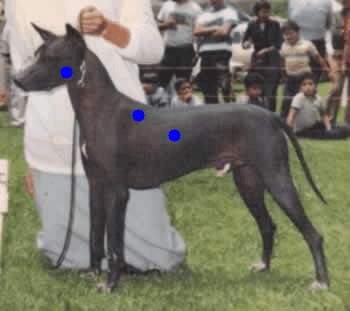} &
\includegraphics[scale=0.25]{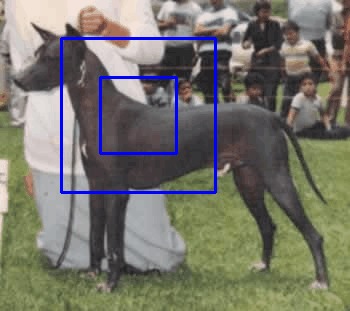} &
\begin{tabular}{c}
\includegraphics[scale=0.25]{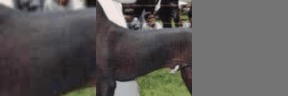} \\
\includegraphics[scale=0.25]{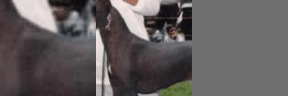} \\
\includegraphics[scale=0.25]{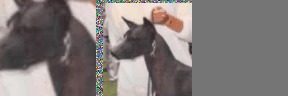}
\end{tabular} &
\includegraphics[scale=0.25]{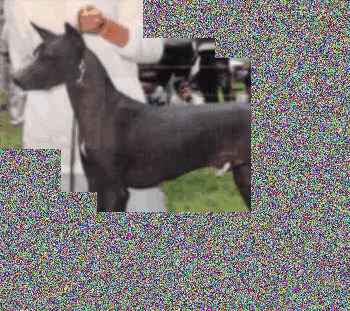} \\
\multicolumn{4}{c}{(a)} \\
\includegraphics[scale=0.25]{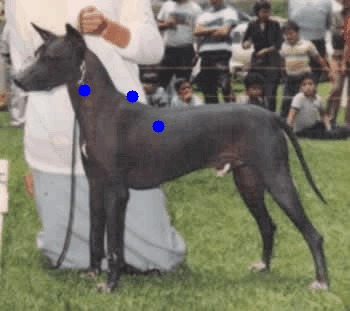} &
\includegraphics[scale=0.25]{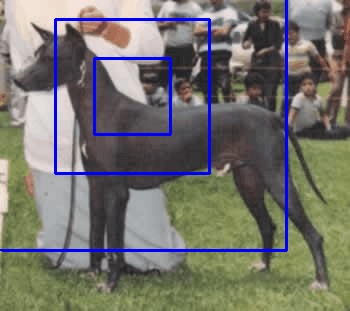} &
\begin{tabular}{c}
\includegraphics[scale=0.25]{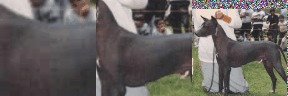} \\
\includegraphics[scale=0.25]{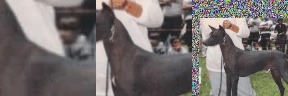} \\
\includegraphics[scale=0.25]{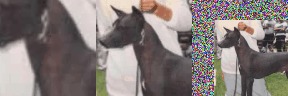}
\end{tabular} &
\includegraphics[scale=0.25]{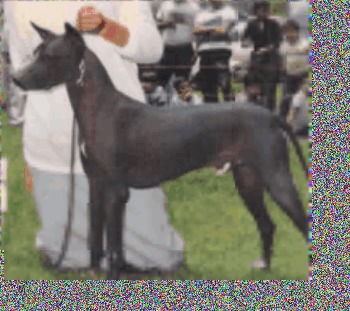} \\
\multicolumn{4}{c}{(b)}
\end{tabular}
\end{center}
\caption{Visualizations of 2-resolution (a) and
3-resolution (b) glimpses on an image
from our validation set, with learned fixation points.
For each the glimpse images are in order, from top to bottom, and
the box diagram corresponds to the second glimpse.
The composite image is created from all three glimpses.
The context image is
not shown but is always the same resolution and size as the low-resolution
glimpse patches shown in (b).}
\label{fig:glimpse}
\end{figure}

The structure of the glimpse images is shown in Figure~\ref{fig:glimpse}.
Each glimpse is a multi-resolution image created by extracting patches
at varying sizes, resizing them all to 96$\times$96, and concatenating them
side-by-side. The goal is to emulate a ``foveal'' structure with the sharpest
part of the image in the center and lower resolution toward the periphery.
The top row shows glimpses for a 2-resolution model and the
bottom row for a 3-resolution model. The high-resolution patch is extracted
from a square that is a fixed size for a given image (more on scale selection
below). The medium-resolution patch is from a square that is twice the
length on a side of the high-resolution patch, and the low-resolution patch
is twice the length of the medium-resolution patch on a side.
For example, if the high resolution patch is 100$\times$100, then the
medium- and low-resolution patches are 200$\times$200 and 400$\times$400,
respectively. Where an extraction box extends off the edge of the image, we
fill the pixels with random noise. Figure~\ref{fig:glimpse} shows composite
images which are a helpful visualization to understand what pixels the network
sees in aggregate, though the network is not presented with them in this form;
these images
are generated from the glimpse pixels by displaying at each pixel the
highest-resolution pixel available in any glimpse, and any pixels not
captured by the glimpses are filled with noise.

\newcolumntype{V}{>{\centering\arraybackslash} m{.2\linewidth} }

The model begins processing from a ``context image'', which is a square
low-resolution patch from the input image that is the same size as our
low-resolution glimpse patch and is also resized to 96$\times$96.
The location of the context image
is chosen randomly in training, but it is centered during inference when it
captures the central square of the image.
The context image is used by layer $r_o^{(2)}$ to produce the first glimpse
location $l_0$ and can influence the selection of subsequent glimpse locations
through the recurrent connections between the $r_n^{(2)}$ layers
along the ``top deck'' of the model. However, the double-decker structure
prevents the context image from having a pathway to
the classifier except through the $l_n$ coordinates. See \cite{BaICLR2015}
for more discussion about this design choice.
In training, the $r_n^{(2)}$ layers that produce the $\hat{l}_n$ coordinates
are trained with a mix of backpropagation from the connection to
layer $r_{n+1}^{(1)}$ and a policy gradient update.

There are four major differences between our system and the
classification-type models from \citet{BaICLR2015}.
First, there is wide variation in image size across our data set, however
the size of the objects scales with the image size most of the time.
To be robust to input image size, our multi-resolution patches are sized
relative to the input image.
In our experiments, a side of the
high-resolution square patch is $1/4$ the shorter dimension of the input image,
making the sides of the medium- and low-resolution patches $1/2$ and the
full length of the image's short side, respectively.

Second, we use a ``vanilla'' RNN instead of an LSTM, where
$r_n^{(1)}$ and $r_n^{(2)}$ at glimpse $n$ each consist
of 4,096 nodes, and $r_{n}{(i)}$ is fully-connected to
$r_{n+1}{(i)}$ for $i=1,2$.
Third, instead of element-wise multiplying
the outputs of the glimpse visual core $G_{image}(x_n|W_{image})$ and
$G_{loc}(l_n|W_{loc})$, our model
linearly combines them
by concatenating their outputs and passing through a fully-connected layer.
Future experiments will incorporate both of these variations.

The final and largest difference is that we replace the
visual glimpse network $G_{image}(x_n|W_{image})$ described in
\cite{BaICLR2015} with a more powerful visual core based on the
``GoogLeNet'' model~\citep{SzegedyArxiv2014} that won the ILSVRC 2014
classification challenge. We start from a GoogLeNet model
that we pre-trained on the ImageNet 1000-way task on a subset of the ILSVRC 2012
training data  \citep{ILSVRC} (more on our use of this data set below).

\begin{figure}
\begin{tabular}{cc}
\includegraphics[width=2.5in]{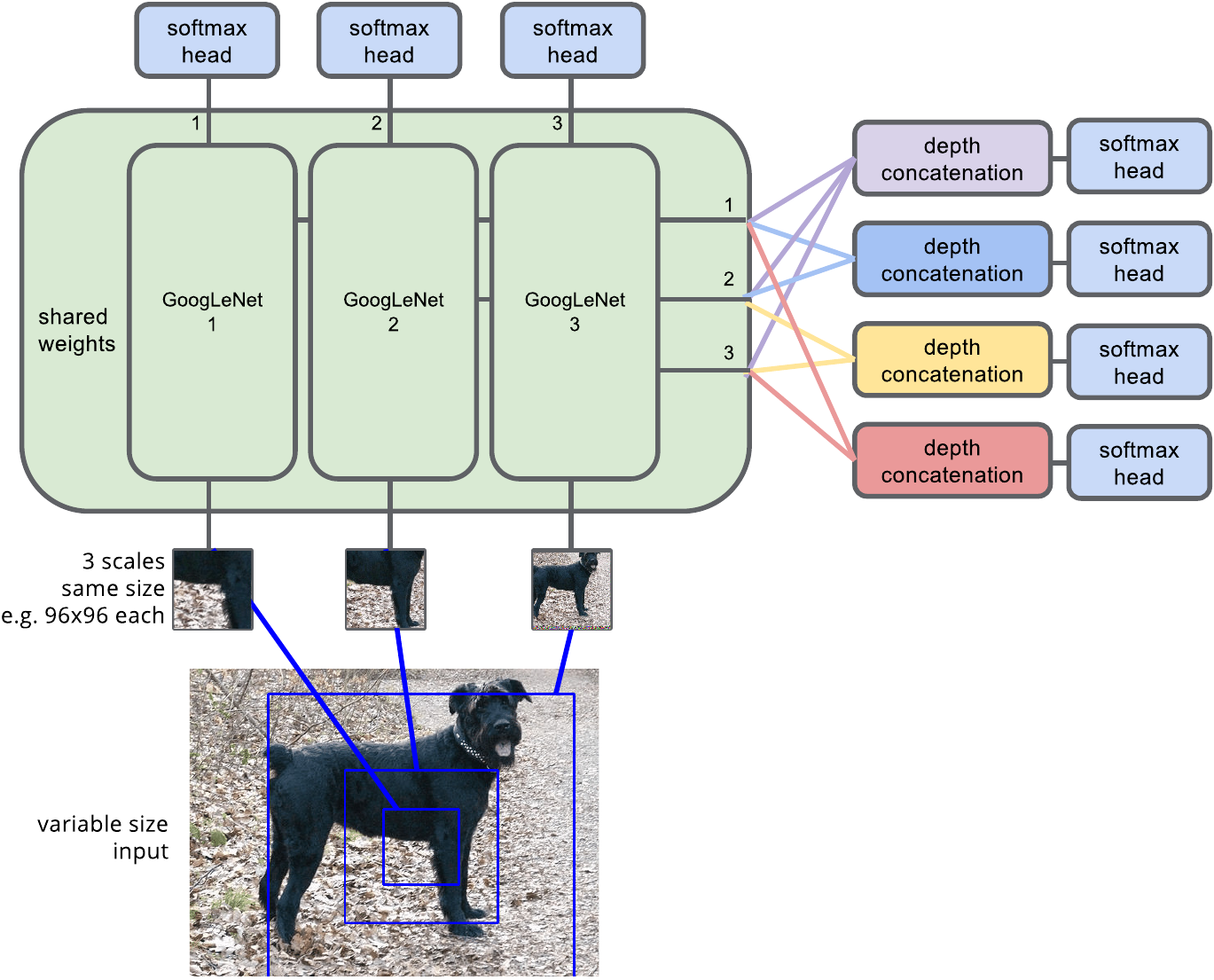} &
\includegraphics[width=2.5in]{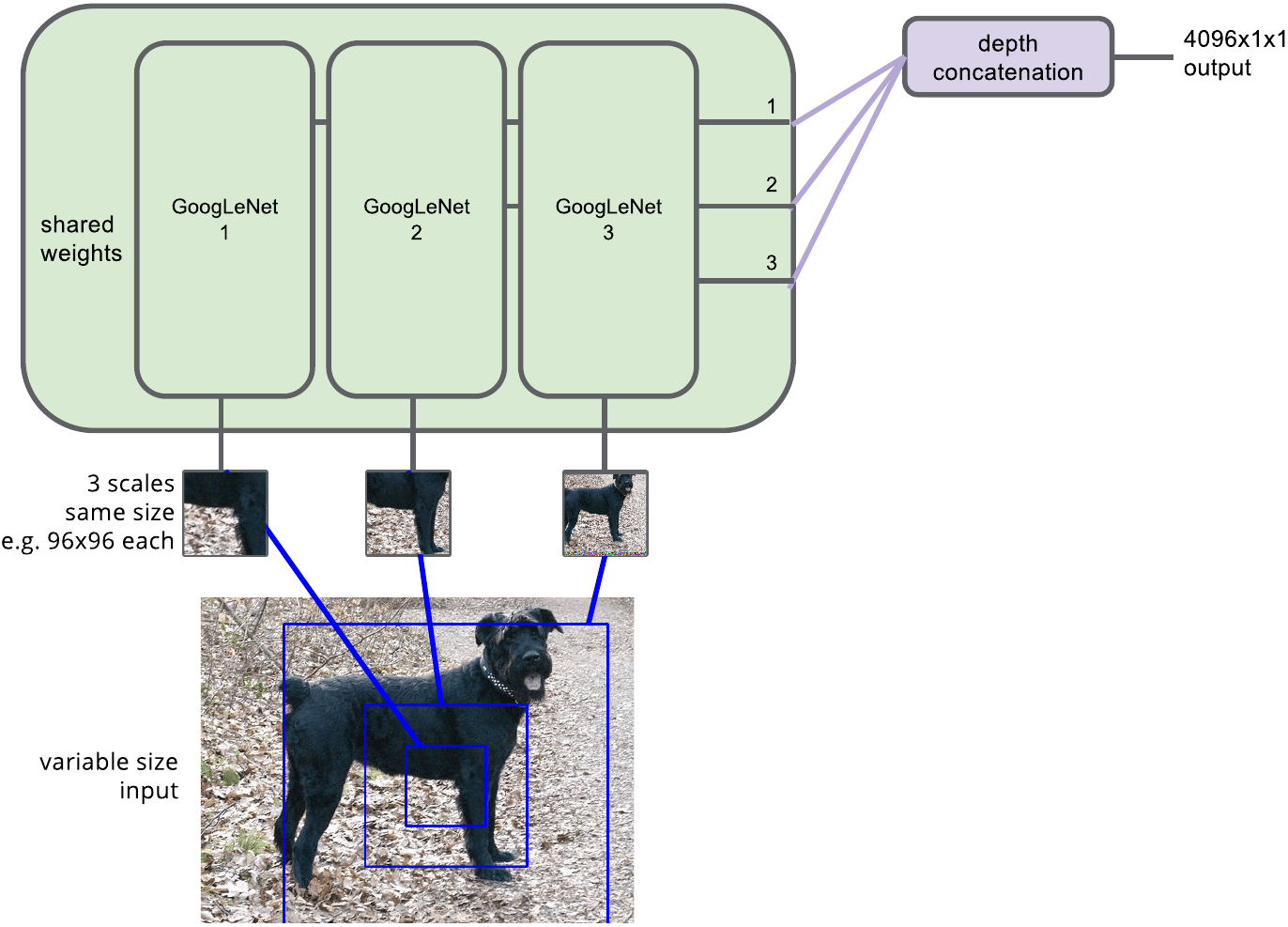}
\end{tabular}
\caption{Pre-training of the visual core (left) and inference within the RNN
(right).}
\label{fig:inception}
\end{figure}

We then fine-tune the visual network outside of the RNN model on
ILSVRC images using random multi-scale patches as input and targeting
the ImageNet 1000-way classification labels. In this stage of training,
we replicate the visual model for each input scale, yielding 3 ``towers''
which share parameters and join their outputs in different combinations
with depth-concatenating layers (Figure \ref{fig:inception}).
All towers are jointly trained by
back-propagating the loss from multiple 1000-way softmax classifiers (called
``heads'') as shown in the figure.
This multi-headed training model ensures each tower remains
independently relevant even if another tower is more informative. We have found
that if taken independently, the lowest-resolution patch typically yields best
results and learning might rely on it solely otherwise.

We initially used a truncated version of GoogLeNet as the visual core of the
RNN because the full model is designed for
224$\times$224 inputs and if applied
to 96$\times$96 inputs, the subsampling and pooling layers cause the final
output to be too small for our purposes. To remedy this, we initially chopped off the
last two ``inception'' layers, skipping five
convolutional layers and an average-pooling layer. We later discovered a large
gain in performance by changing the stride of the first
convolution of the network from two to one and restoring the visual network to its
full depth. Stride-1 convolutions were historically
used in early deep learning works, however \cite{krizhevsky2012imagenet} later
popularized strides of two in early layers for efficiency reasons.
In our experiments, we changed the stride only in the fine-tuning phase, starting
from a pre-trained model with a stride of two.
Although it is not obvious that changing the stride of a pre-trained model should
work, when incorporated with the RNN, the best performance of the
three-resolution, three-glimpse attention model increased from
68\% to 76.8\%.

During training of the attention model, we remove all training
heads and take the output of the depth concatenation of multiple towers as
glimpse input as shown in Figure~\ref{fig:inception}. For this work,
we hold the visual core's parameters fixed during attention model training.
We pre-train with all three resolutions, but when used in the RNN,
we vary across experiments which subsets of the resolutions are used, so
the training regime in those experiments is slightly different than testing.

In all stages of training for the attention RNN and our experimental baselines,
we use a subset of the ILSVRC 2012 training set from which we removed the
Stanford Dogs test images as well as the Dogs images that we use for
validation. We refer to this in our experimental section as the ``de-duped''
ILSVRC data set. Pre-training with de-duped data instead of the original set
does make a small difference in performance: we saw a drop of 3\% accuracy in the
full GoogLeNet baseline model when trained with de-duped data relative to one
trained with the full ILSVRC 2012 training set.

Finally, it is worth noting that when fine-tuning the visual core, we did not use
the Stanford Dogs training set, and since the parameters of the visual core
are held fixed while training the RNN on Dogs, this means the powerful
visual processing component in the RNN is not trained on the final task.
We performed an experiment with a visual core fine-tuned on the Stanford
Dogs training data, and we did not see an increase in performance, demonstrating
again that the final RNN model is fairly robust to the pre-training and
fine-tuning procedure.


\section{Experimental Results}
\label{sec:results}

We trained and evaluated our model on the Stanford Dogs fine-grained
categorization data set~\citep{Dogs2011}.
The task is to
categorize each of 8,580 test images as containing one of 120 dog
breeds. The training set consists of 100 images per class, and the
test images are unevenly distributed across classes, averaging about 71 test
images per class. The training and test sets both include bounding boxes
that provide a tight crop around the target dog, and while the best results
we know of
in the literature use the bounding boxes both in
training and testing, we use neither the training nor testing boxes.
We follow the practice common in the literature of
augmenting the training set by reflecting the images along the vertical axis.
Our model starts from the full images, without cropping, reshaping, or scaling.
We performed experiments and chose hyperparameters using an 80/20
training/validation split of the Stanford Dogs training set. We selected
hyperparameters (e.g. learning rate, sample variance) using the
20\% validation set then trained on the full training set, and we
only performed final evaluation on the Dogs test set on our selected
hyperparameters.

The background in the images is not highly correlated with the class label,
so any method not using the bounding boxes needs to localize the object of
interest in order to classify it.
This is a nice task to explore for our attention model in a couple ways:
(1) the model can use the context image in order to focus its
glimpses on the object of interest, and (2) we can intuit which parts of the
image the model should observe to make a prediction. With many
other natural image object classification data sets, such as ImageNet, the
signal from the surrounding context is mixed with the object for classification
(e.g. boats are expected to be surrounded by water). The size of the data
set is also more suitable to a deep learning method than most other available
fine-grained data sets, though Caltech-UCSD Birds 2011~\citep{WahCUB_200_2011}
is similar in size with 12,000 training images for 200 categories.\footnote{We
have not yet experimented with Caltech-UCSD Birds 2011, but the approach here should
apply nicely.}
Lastly, it remains a difficult data set, with a large amount of
intra-class variation, similarity across
classes, and large variation in pose, lighting, and background
(Figure~\ref{fig:dogsAreHard}).

Table~\ref{tab:numbers}(a) shows the mean accuracy (mA) for
different combinations of resolutions and number of glimpses.
We experimented with high, medium, and low resolutions individually,
medium and high combined, and all three resolutions and with
one, two, and three glimpses. The table also shows
previously published results on the data set\footnote{Missing from the results
are entries to the FGComp 2013 fine-grained competition. There are
high-performing entries
from deep learning models in the dogs category, though to our knowledge these
models have
not been published. {\em CognitiveVision} and {\em CafeNet} scored 61\%
and 57\% on the challenge, respectively, using bounding boxes both in
training and testing. The challenge training set is from Stanford Dogs, but
the test set
is independent, the class labels have not been made public, and the
evaluation server is no longer running.
As such, we cannot compare directly to these numbers,
but we have been told anecdotally that scores on the
FGComp 2013 challenge tend to be about 10\% absolute lower than on the
Stanford Dogs test set.}.
Versions of our model that use medium and low resolution patches
outperform state-of-the-art results.
When using only two small high-resolution patches from the image,
our model matches the best published result.
All previously published results shown use ground truth bounding boxes
for training and testing while our model does not.


\begin{table}
\caption{Results on Stanford Dogs for (a) our RNN model and (b) our
GoogLeNet baselines and previous state-of-the-art results,
measured by mean accuracy percentage (mA) as described in
\citet{ChaiICCV2013}. The GoogLeNet baseline models were pre-trained on the
de-duped ILSVRC 2012 training set and fine-tuned with the Stanford Dogs
training set. Results marked with a star indicate use of tight
ground truth bounding boxes around the dogs in training and testing.}
\label{tab:numbers}
\begin{center}
\begin{tabular}{cc}
\begin{tabular}{lccc}
\# glimpses & 1 & 2 & 3 \\
\hline
high res only & 43.5 & 48.3 & 49.6 \\
medium res only & 70.1 & 72.3 & 72.8 \\
low res only & 70.3 & 70.1 & 70.7 \\
high+medium res & 70.7 & 72.6 & 72.7 \\
3-resolution & 76.3 & 76.5 & \textbf{76.8}
\end{tabular} &
\begin{tabular}{ll}
\\
\citet{YangNIPS2012}* & 38.0 \\
\citet{ChaiICCV2013}* & 45.6 \\
\citet{GavvesICCV2013}* & 50.1 \\
GoogLeNet 96$\times$96 & 58.8 \\
GoogLeNet 224$\times$224 & \textbf{75.5}
\end{tabular} \\
(a) & (b)
\end{tabular}
\end{center}
\end{table}


\begin{figure}
\begin{center}
\fcolorbox{green}{green}{
\includegraphics[width=0.26\linewidth]{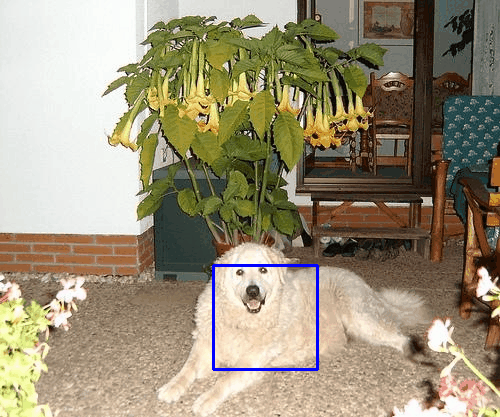}
\includegraphics[width=0.06\linewidth]{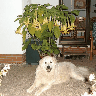}
\includegraphics[width=0.26\linewidth]{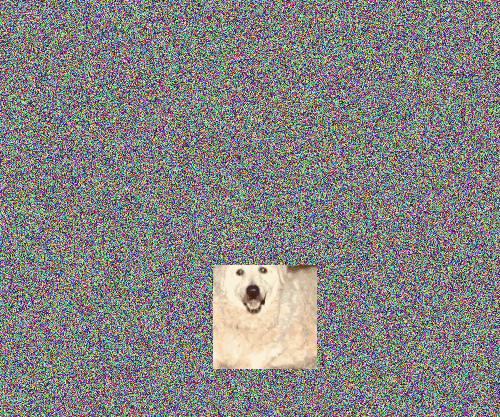}}
\\
\fcolorbox{green}{green}{
\includegraphics[width=0.26\linewidth]{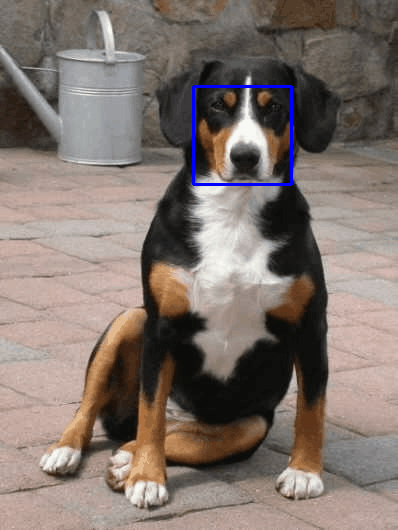}
\includegraphics[width=0.06\linewidth]{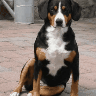}
\includegraphics[width=0.26\linewidth]{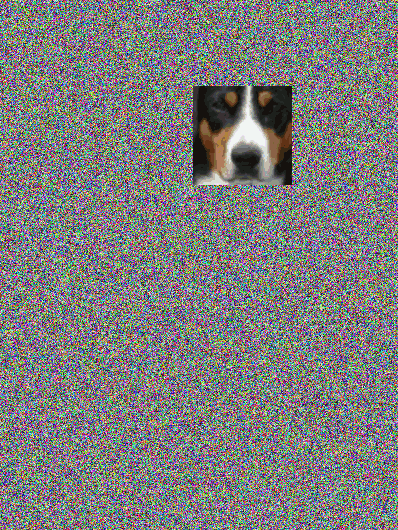}}
\\
\fcolorbox{red}{red}{
\includegraphics[width=0.26\linewidth]{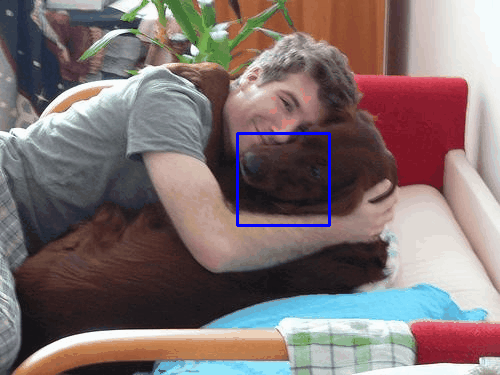}
\includegraphics[width=0.06\linewidth]{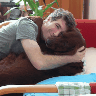}
\includegraphics[width=0.26\linewidth]{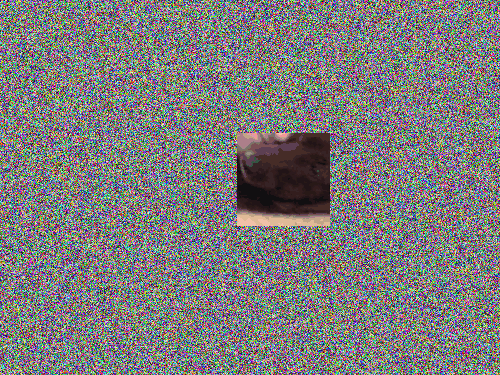}} \\
\begin{tabular}{>{\centering}m{0.26\linewidth} >{\centering}m{0.06\linewidth} >{\centering}m{0.26\linewidth} }
(a) & (b) & (c)
\end{tabular}
\caption{Selected examples of the high-resolution, one-glimpse model run on the validation set. The system takes (a) the original image, subsamples it to create (b) the context image, and uses the context to select (c) a single high resolution glimpse. The outline of the high-resolution glimpse is also shown on the full image in (a) for context. Rows surrounded by green were correctly classified, while red indicates a classification error. Note that while the bottom example is misclassified, the model still learned to look at the dog face despite clutter, occlusion, and an uncommon pose. See Figure \ref{fig:random_validation_samples} for a random selection of results.}
\label{fig:curated_validation_samples}
  \end{center}
\end{figure}

While the high-resolution single-glimpse model has the lowest performance of
the set, visualizations of the selected glimpse locations show that it is
learning to take a good first action (Figure~\ref{fig:curated_validation_samples}).
These are fairly representative of the behavior of the model, which most
frequently chooses a patch on or near the dog's face. While it may make an
informative first glimpse, it is often not
able to correctly classify the dog from that single sample.
It is important to note that the model automatically learned to focus on the most
discriminative features such as faces and fur without ever receiving spatial clues
such as bounding boxes. This is pretty remarkable in that bounding boxes
are usually required for good performance on this task,
and obtaining bounding boxes at scale is difficult and expensive. It also
raises the possibility of attention models providing a signal for detection without
labeled bounding boxes.
The figure shows
two images where it classified the dog correctly in green, and one in red
where it assigned an incorrect label. One pathological pattern we noticed is
that if there are two dogs in the image, it often chooses a patch that is
halfway between the two, which is likely due to the regression-style
output of the glimpse coordinates which may encourage the model to output the
average of two predicted targets. A set of randomly chosen examples is shown
in the appendix
with Figure~\ref{fig:random_validation_samples}.

Comparisons to results not using deep learning do not give a good sense of the
strength of the model, however.
In the last couple years deep nets have been winning
the ILSVRC classification challenge by a significant margin, so it may be
expected that a deep neural net would outperform the existing results.
To address this we also evaluated GoogLeNet on the full image without the attention
RNN.
We experimented with two baseline versions of GoogLeNet: ``full'' and
``low-resolution''. The full GoogLeNet uses the same architecture as in
\citet{SzegedyArxiv2014} and is trained and tested on 224$\times$224 padded
versions of the full Dog images. This is the strongest of the two baselines.
The low-resolution GoogLeNet has the same architecture as our
RNN visual core and also takes 96$\times$96 inputs and uses a stride of one for
the first convolution. It does not have three
input scales like the RNN visual core but instead takes the full image with
padding that centers it. The low-resolution GoogLeNet input is close in resolution
to the low-resolution foveal input to our attention model.
Both versions were pre-trained using
the de-duped ILSVRC 2012 data set,
then the top fully-connected layer and softmax layers were resized and
reinitialized, and the full model was trained to convergence on the
Dogs training set. In addition to the mirroring applied for RNN training,
brightness and color transformations were also applied to the training images
for the baselines. Unlike \citet{SzegedyArxiv2014}, for this comparison we did
not average across patches or different GoogLeNet models.

Our three-resolution, one-glimpse attention model reached 76.3\% mA compared
to 75.5\% for the full GoogLeNet model (\ref{tab:numbers}(b)). This version of
the attention model gets three 96$\times$96 inputs it can use for classification and
the 96$\times$96 context image for choosing the position of the one glimpse.
Compared to the 224$\times$224 input to GoogLeNet, we perform better with
73\% of the pixels. This doesn't reduce the amount of computation, however,
because the attention model uses a stride of one in the first
convolutional layer compared to two in the full GoogLeNet.
With three glimpses, the performance increases slightly to 76.8\%, though this
almost triples the number of pixels input\footnote{For a fairer comparison
between multiple-glimpse attention and GoogLeNet, we would ideally
take an equal number of random crops as input to GoogLeNet and average
the outputs.}. With only medium and high resolution inputs (72.7\%,
three glimpses), the attention model is nearing the performance of
full GoogLeNet with 55\% of the pixels.

A comparison between the low-resolution GoogLeNet (58.8\%) and our low resolution
attention model (70.3\%, one glimpse) shows that the increase in convolution stride
does not account for the strong performance because both models share the same visual
network and similar resolution inputs (depending on ratio of the long and short
sides of the image).
This indicates the large difference in performance is due to the network
choosing an informative part of the image to use as input for classification.

Lastly, it is interesting to compare one-, two-, and three- glimpse results
for different resolution inputs. Using three resolutions, the performance
only increases slightly from 76.3\% for one glimpse to 76.8\% for three. One likely
cause is that the three-resolution glimpse contains enough information from
the full image that the information gained from additional glimpses is minimal.
An additional piece of evidence is that the performance order of low- and medium-only
resolution models swap when going from one to two glimpses.
We ran the high resolution-only glimpse experiment to test this; the results
for one, two, and three glimpses are 43.5\%, 48.3\%, and 49.6\%, respectively,
demonstrating that when the amount of information in each glimpse is restricted,
the model benefits more from several glimpses. However, the
improvement with increased number of glimpses flattens quickly, indicating that the
model has limited capacity to make use of more than two or three glimpses. One
hypothesis is the RNN is not passing enough of the information from the early
glimpses along to the classification layer. It is future work to explore using
LSTM cells and increasing the recurrent capacity of the network.



\bibliography{iclr2015}
\bibliographystyle{iclr2015}

\appendix
\section{Random validation samples}

\begin{figure}[h!]
\setlength{\tabcolsep}{1pt}
\renewcommand{\arraystretch}{1.5}
\begin{tabular}{cccc}
\parbox[b][][c]{0.2\linewidth}{\centering 3 resolutions,\\3 glimpses} &
\parbox[b][][c]{0.2\linewidth}{\centering 3 resolutions,\\1 glimpse} &
\parbox[b][][c]{0.2\linewidth}{\centering high-res only,\\3 glimpses} &
\parbox[b][][c]{0.2\linewidth}{\centering high-res only,\\1 glimpse}
\\
\fcolorbox{green}{green}{
\includegraphics[width=0.104\linewidth]{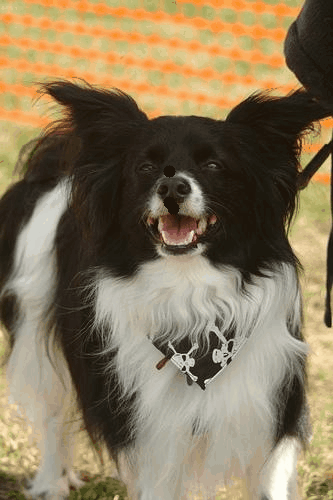}
\includegraphics[width=0.104\linewidth]{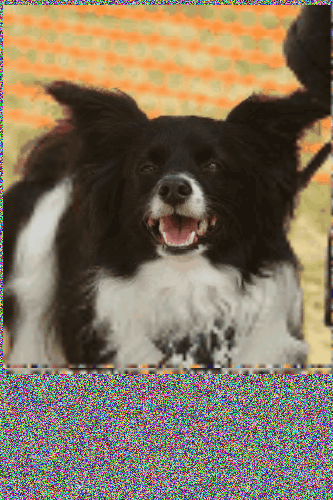}}
& \fcolorbox{green}{green}{
\includegraphics[width=0.104\linewidth]{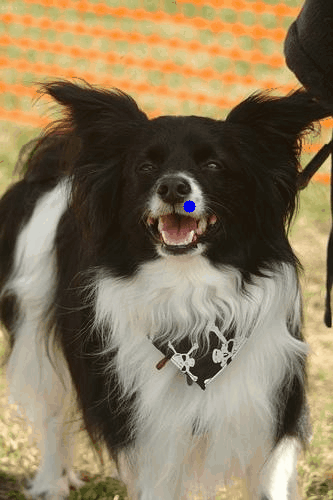}
\includegraphics[width=0.104\linewidth]{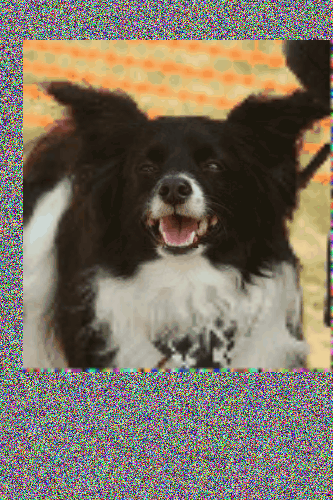}}
& \fcolorbox{red}{red}{
\includegraphics[width=0.104\linewidth]{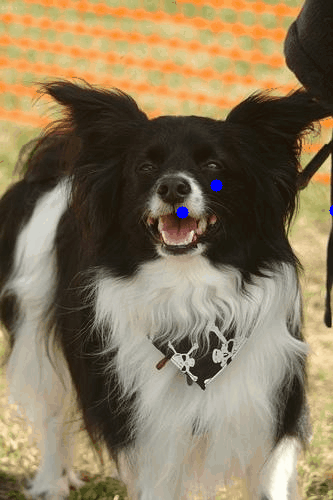}
\includegraphics[width=0.104\linewidth]{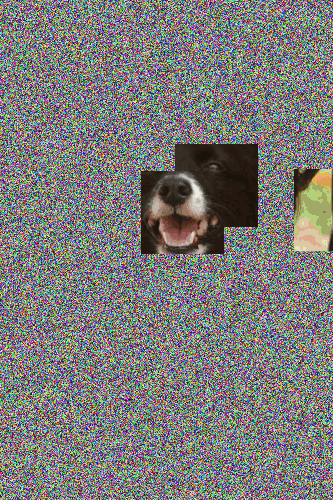}}
& \fcolorbox{red}{red}{
\includegraphics[width=0.104\linewidth]{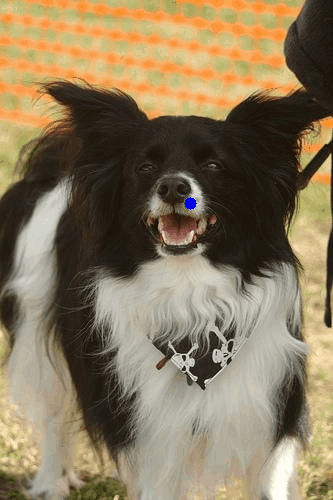}
\includegraphics[width=0.104\linewidth]{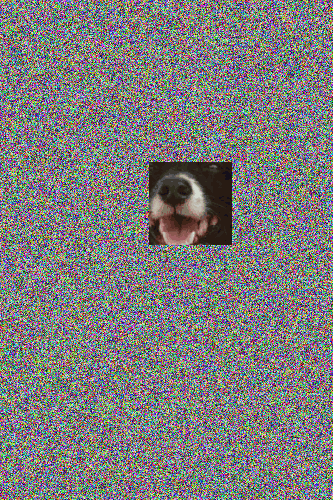}}
\\
\begin{tabular}{cc}
\parbox[b][][c]{0.104\linewidth}{\centering original} &
\parbox[b][][c]{0.104\linewidth}{\centering composite}
\end{tabular} &
\begin{tabular}{cc}
\parbox[b][][c]{0.104\linewidth}{\centering original} &
\parbox[b][][c]{0.104\linewidth}{\centering composite}
\end{tabular} &
\begin{tabular}{cc}
\parbox[b][][c]{0.104\linewidth}{\centering original} &
\parbox[b][][c]{0.104\linewidth}{\centering composite}
\end{tabular} &
\begin{tabular}{cc}
\parbox[b][][c]{0.104\linewidth}{\centering original} &
\parbox[b][][c]{0.104\linewidth}{\centering composite}
\end{tabular}
\\
\fcolorbox{red}{red}{
\includegraphics[width=0.104\linewidth]{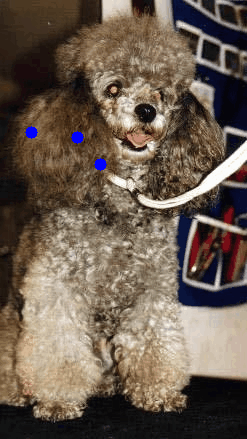}
\includegraphics[width=0.104\linewidth]{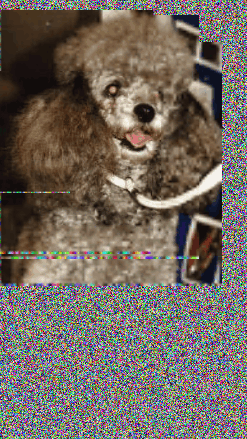}}
& \fcolorbox{red}{red}{
\includegraphics[width=0.104\linewidth]{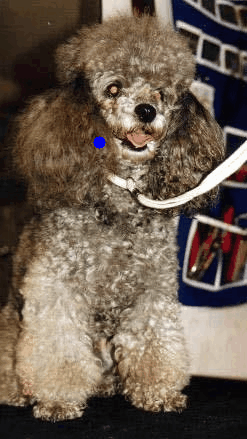}
\includegraphics[width=0.104\linewidth]{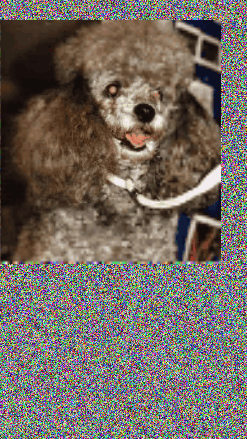}}
& \fcolorbox{red}{red}{
\includegraphics[width=0.104\linewidth]{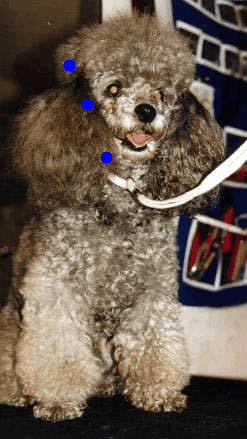}
\includegraphics[width=0.104\linewidth]{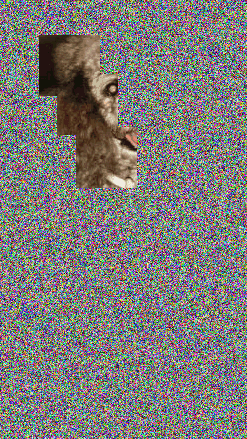}}
& \fcolorbox{red}{red}{
\includegraphics[width=0.104\linewidth]{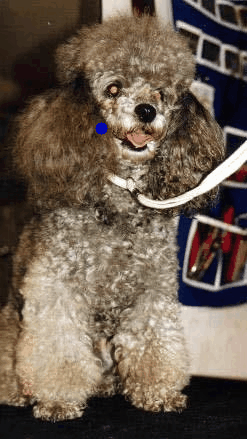}
\includegraphics[width=0.104\linewidth]{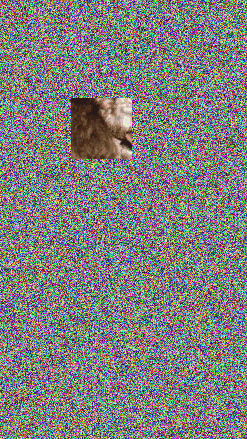}}
\\ 
\fcolorbox{green}{green}{
\includegraphics[width=0.104\linewidth]{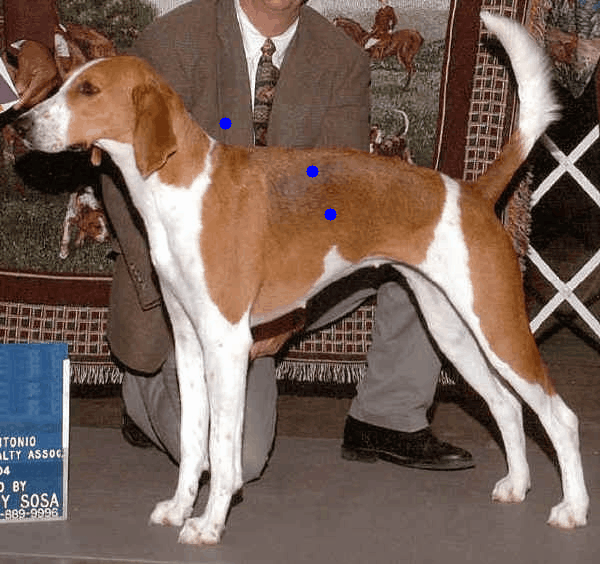}
\includegraphics[width=0.104\linewidth]{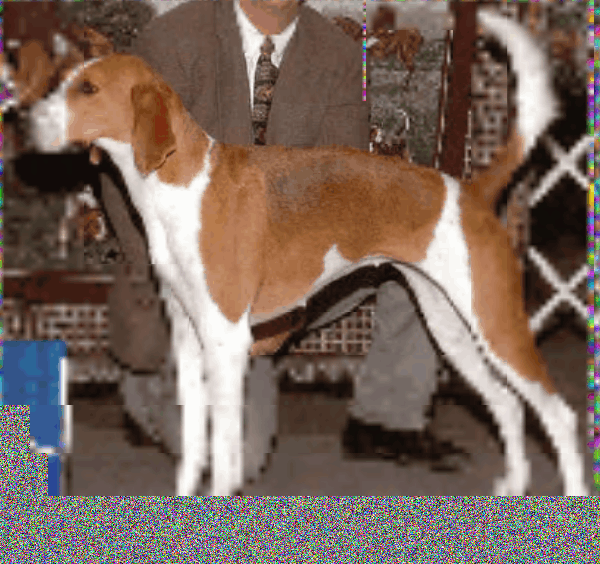}}
& \fcolorbox{green}{green}{
\includegraphics[width=0.104\linewidth]{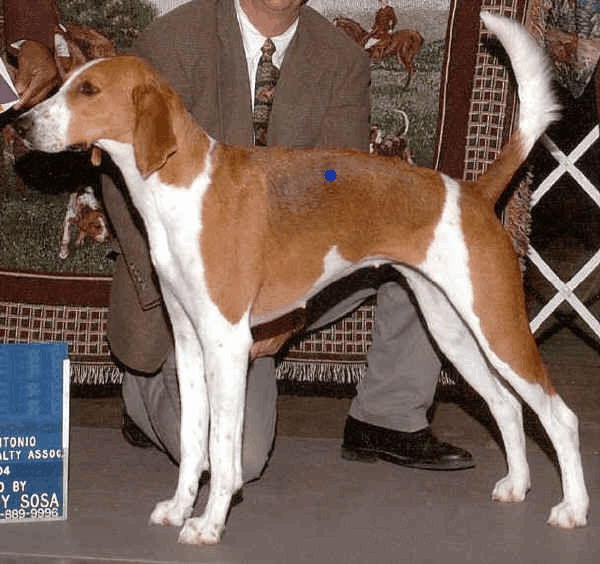}
\includegraphics[width=0.104\linewidth]{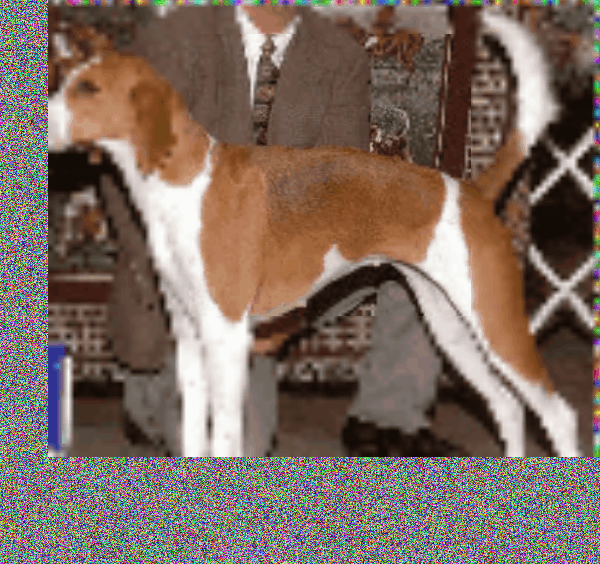}}
& \fcolorbox{red}{red}{
\includegraphics[width=0.104\linewidth]{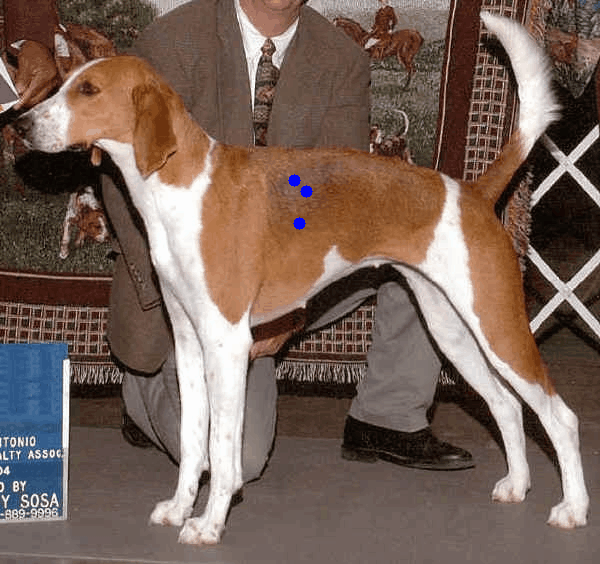}
\includegraphics[width=0.104\linewidth]{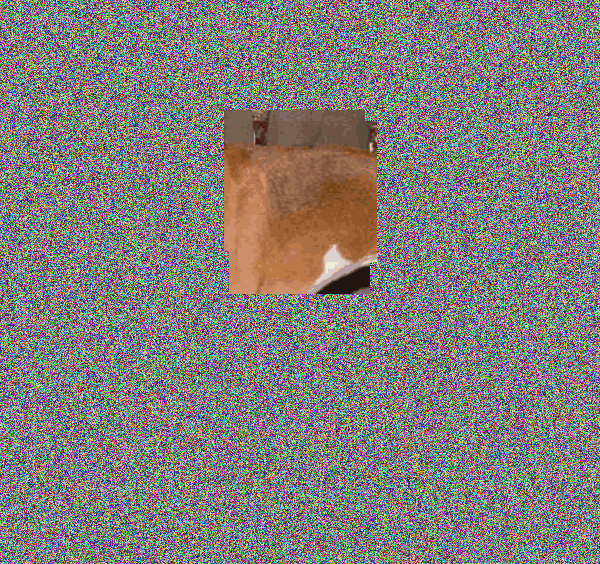}}
& \fcolorbox{red}{red}{
\includegraphics[width=0.104\linewidth]{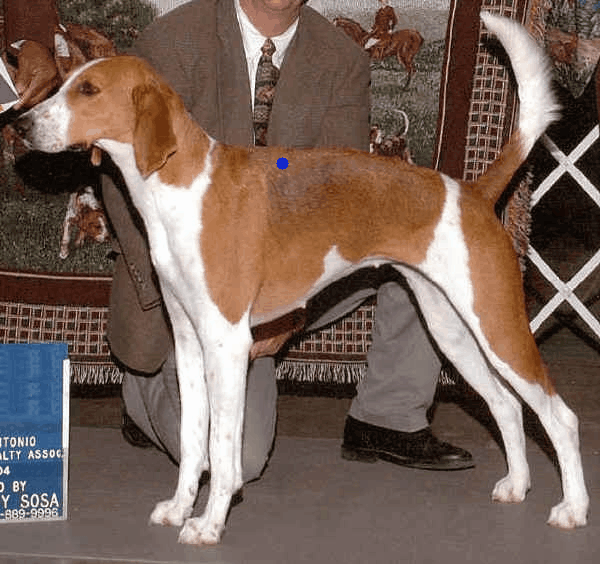}
\includegraphics[width=0.104\linewidth]{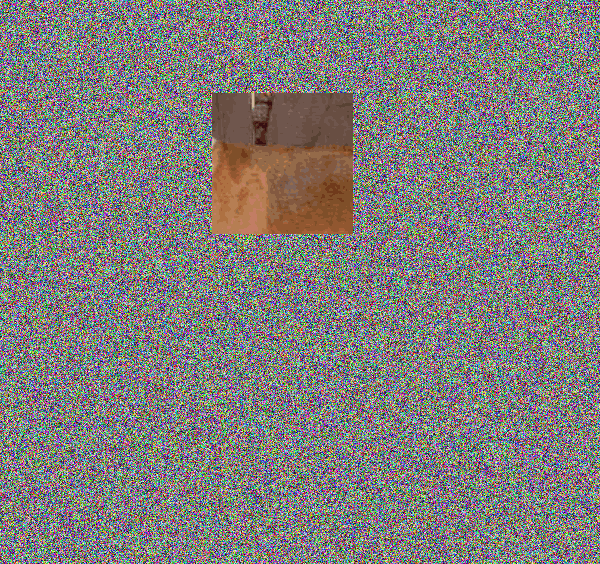}}
\\ 
\fcolorbox{green}{green}{
\includegraphics[width=0.104\linewidth]{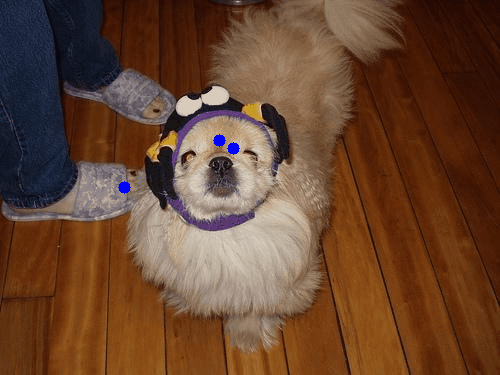}
\includegraphics[width=0.104\linewidth]{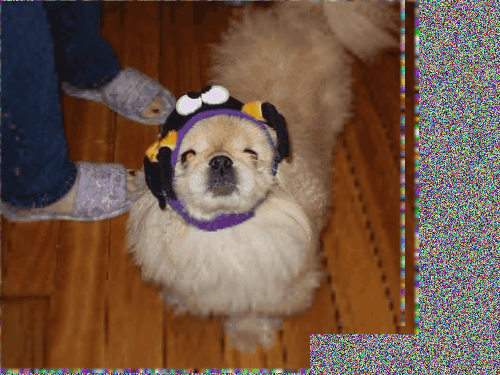}}
& \fcolorbox{green}{green}{
\includegraphics[width=0.104\linewidth]{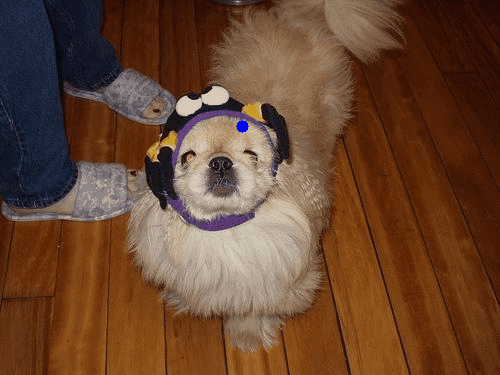}
\includegraphics[width=0.104\linewidth]{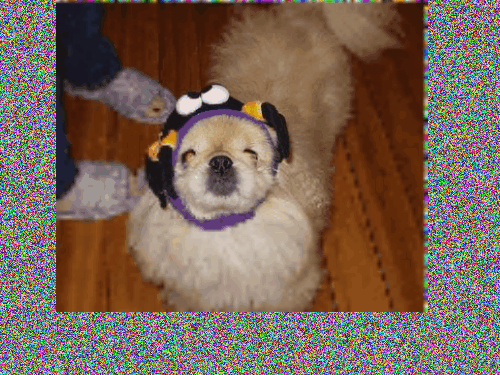}}
& \fcolorbox{green}{green}{
\includegraphics[width=0.104\linewidth]{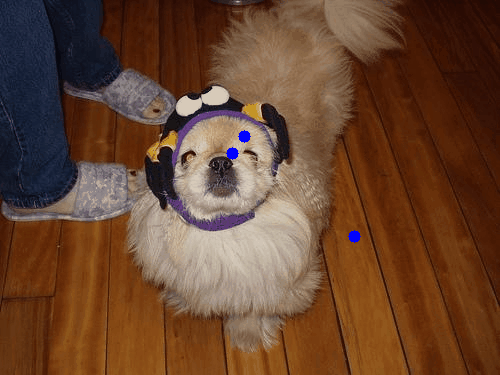}
\includegraphics[width=0.104\linewidth]{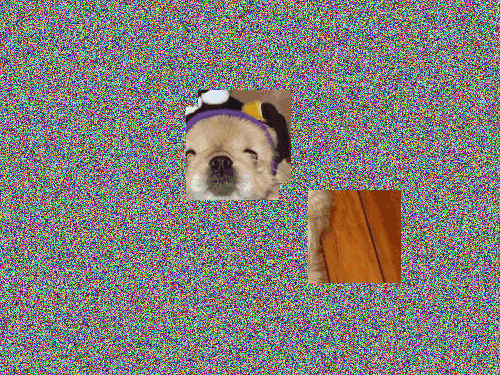}}
& \fcolorbox{green}{green}{
\includegraphics[width=0.104\linewidth]{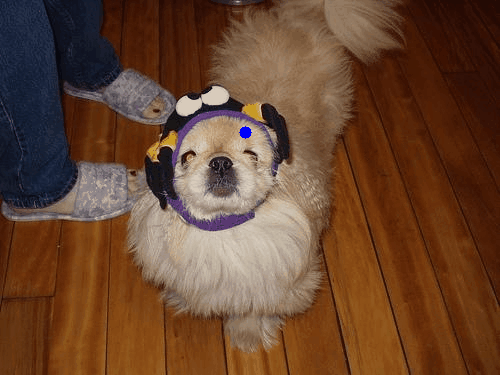}
\includegraphics[width=0.104\linewidth]{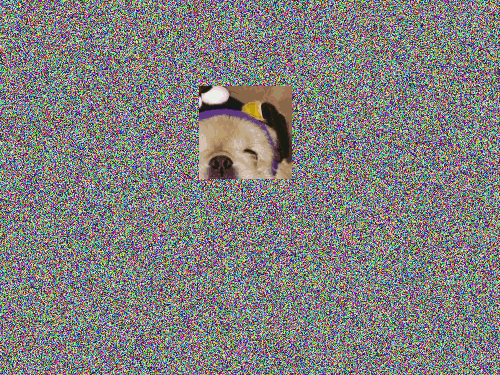}}
\\ 
\fcolorbox{green}{green}{
\includegraphics[width=0.104\linewidth]{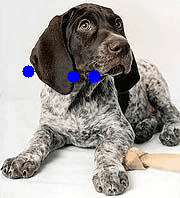}
\includegraphics[width=0.104\linewidth]{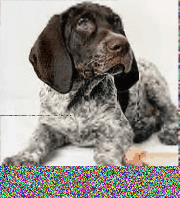}}
& \fcolorbox{green}{green}{
\includegraphics[width=0.104\linewidth]{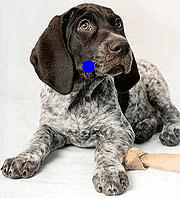}
\includegraphics[width=0.104\linewidth]{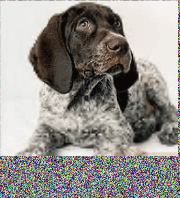}}
& \fcolorbox{green}{green}{
\includegraphics[width=0.104\linewidth]{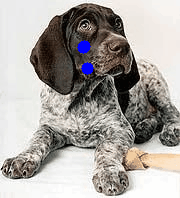}
\includegraphics[width=0.104\linewidth]{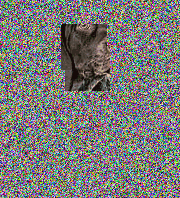}}
& \fcolorbox{red}{red}{
\includegraphics[width=0.104\linewidth]{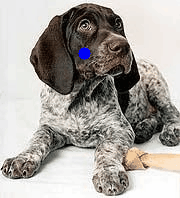}
\includegraphics[width=0.104\linewidth]{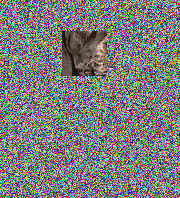}}
\\
%
\fcolorbox{red}{red}{
\includegraphics[width=0.104\linewidth]{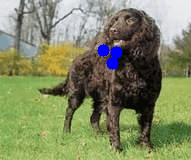}
\includegraphics[width=0.104\linewidth]{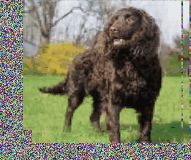}}
& \fcolorbox{red}{red}{
\includegraphics[width=0.104\linewidth]{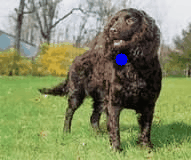}
\includegraphics[width=0.104\linewidth]{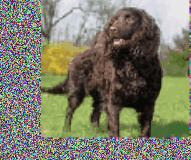}}
& \fcolorbox{red}{red}{
\includegraphics[width=0.104\linewidth]{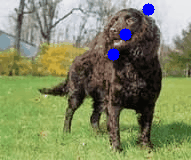}
\includegraphics[width=0.104\linewidth]{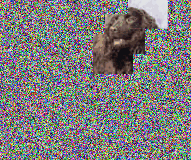}}
& \fcolorbox{red}{red}{
\includegraphics[width=0.104\linewidth]{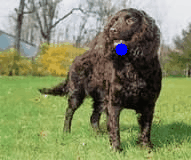}
\includegraphics[width=0.104\linewidth]{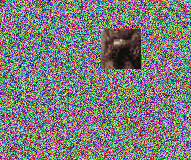}}
\\ 
\fcolorbox{red}{red}{
\includegraphics[width=0.104\linewidth]{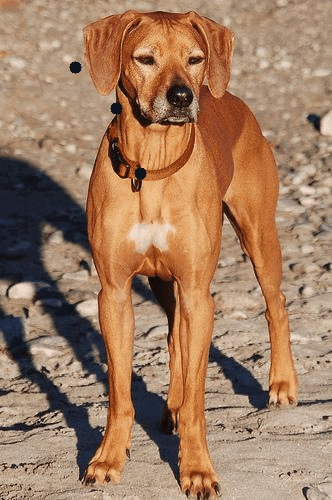}
\includegraphics[width=0.104\linewidth]{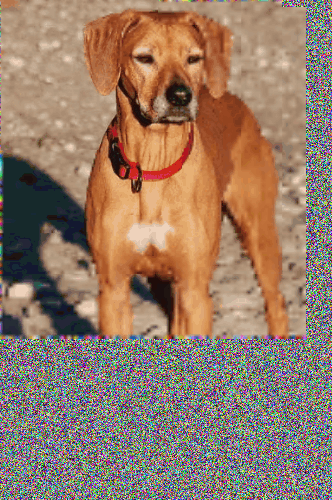}}
& \fcolorbox{red}{red}{
\includegraphics[width=0.104\linewidth]{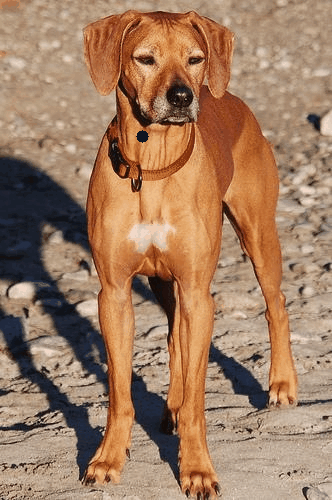}
\includegraphics[width=0.104\linewidth]{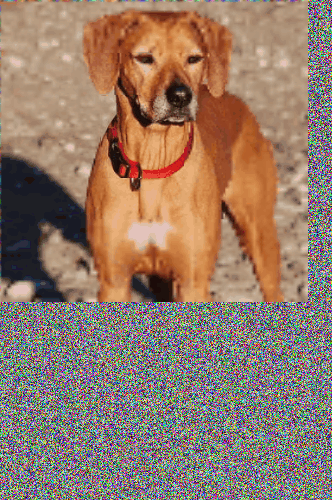}}
& \fcolorbox{red}{red}{
\includegraphics[width=0.104\linewidth]{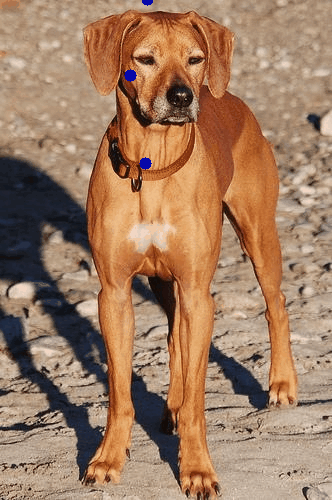}
\includegraphics[width=0.104\linewidth]{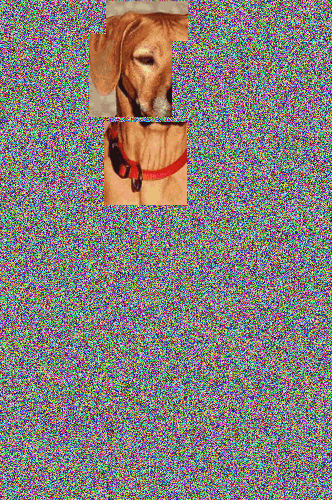}}
& \fcolorbox{red}{red}{
\includegraphics[width=0.104\linewidth]{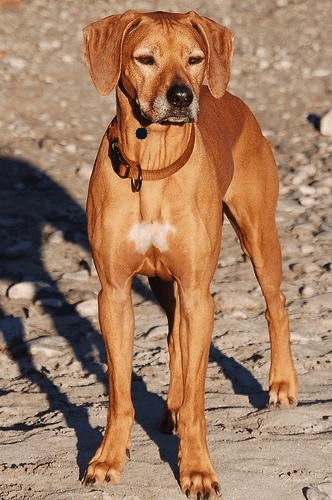}
\includegraphics[width=0.104\linewidth]{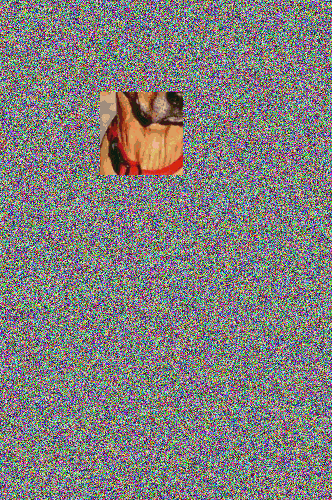}}
\\ 
\fcolorbox{green}{green}{
\includegraphics[width=0.104\linewidth]{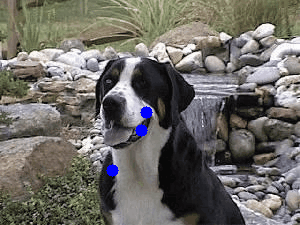}
\includegraphics[width=0.104\linewidth]{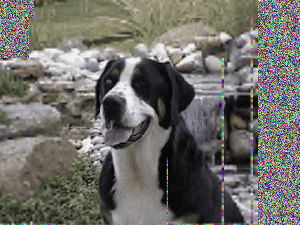}}
& \fcolorbox{green}{green}{
\includegraphics[width=0.104\linewidth]{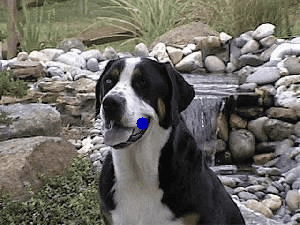}
\includegraphics[width=0.104\linewidth]{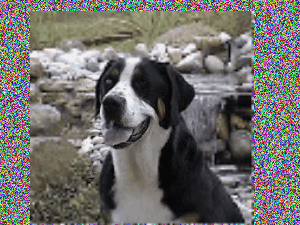}}
& \fcolorbox{red}{red}{
\includegraphics[width=0.104\linewidth]{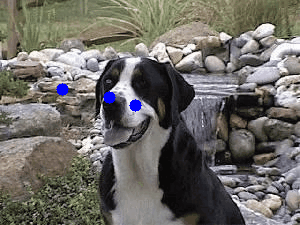}
\includegraphics[width=0.104\linewidth]{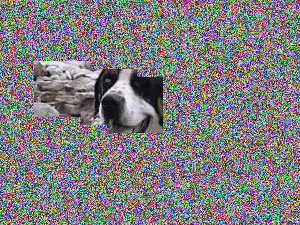}}
& \fcolorbox{red}{red}{
\includegraphics[width=0.104\linewidth]{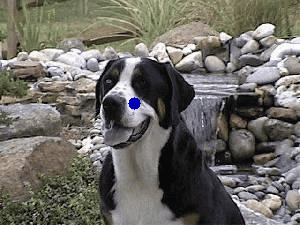}
\includegraphics[width=0.104\linewidth]{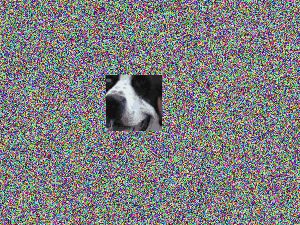}}
\\ 
\fcolorbox{red}{red}{
\includegraphics[width=0.104\linewidth]{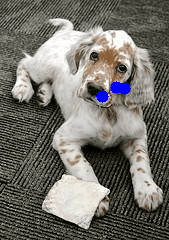}
\includegraphics[width=0.104\linewidth]{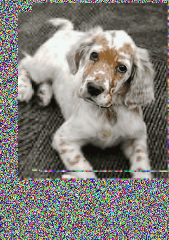}}
& \fcolorbox{red}{red}{
\includegraphics[width=0.104\linewidth]{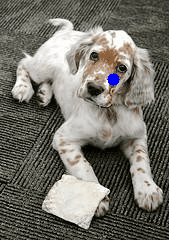}
\includegraphics[width=0.104\linewidth]{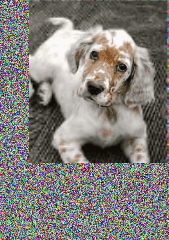}}
& \fcolorbox{green}{green}{
\includegraphics[width=0.104\linewidth]{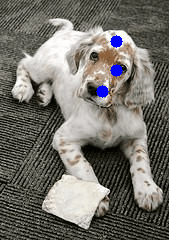}
\includegraphics[width=0.104\linewidth]{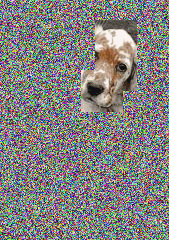}}
& \fcolorbox{red}{red}{
\includegraphics[width=0.104\linewidth]{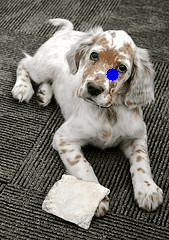}
\includegraphics[width=0.104\linewidth]{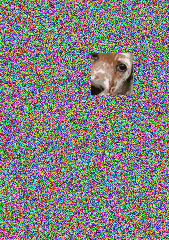}}
\\ 
\fcolorbox{green}{green}{
\includegraphics[width=0.104\linewidth]{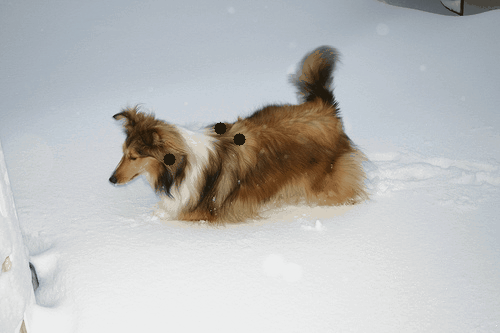}
\includegraphics[width=0.104\linewidth]{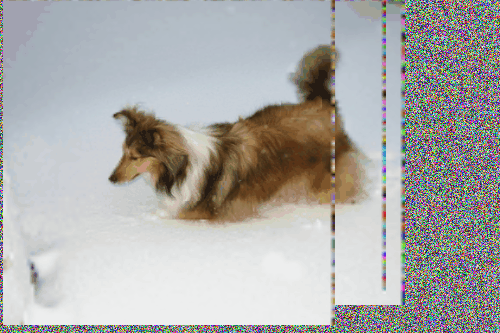}}
& \fcolorbox{green}{green}{
\includegraphics[width=0.104\linewidth]{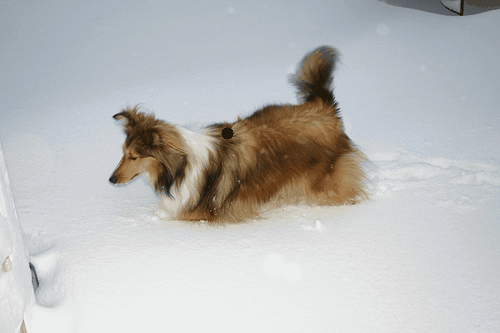}
\includegraphics[width=0.104\linewidth]{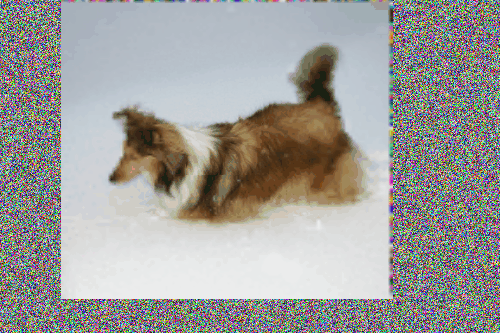}}
& \fcolorbox{green}{green}{
\includegraphics[width=0.104\linewidth]{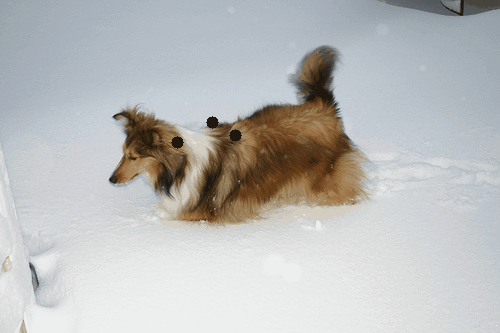}
\includegraphics[width=0.104\linewidth]{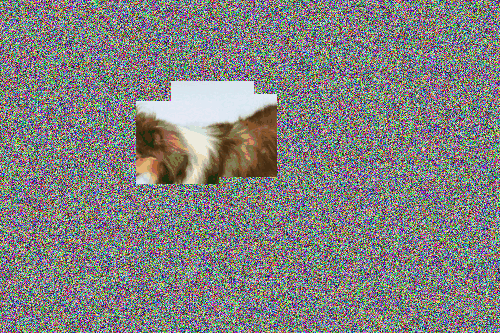}}
& \fcolorbox{green}{green}{
\includegraphics[width=0.104\linewidth]{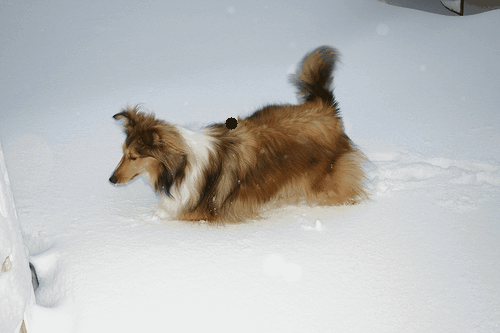}
\includegraphics[width=0.104\linewidth]{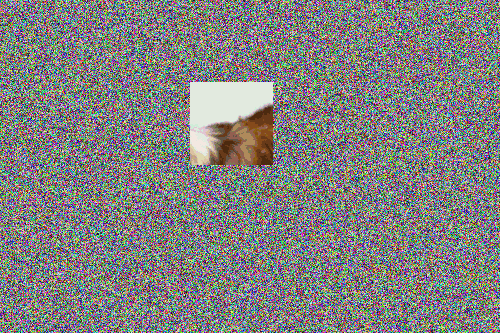}}

\end{tabular}
\caption{Ten randomly-selected validation samples from four different model variations. See text for details.}
\label{fig:random_validation_samples}
\end{figure}

Figure \ref{fig:random_validation_samples} shows ten randomly-selected validation examples from four
different versions of the attention model: three-resolution (low, medium, high) with three glimpses,
three-resolution with one glimpse, high-resolution only with three glimpses, and high-resolution with
one glimpse. For each model and input, on the left it shows the original image with dots at the centers of the
glimpses, and on the right it shows a composite image. Green borders
indicate a correct classification, and red borders indicate an error.
The leftmost column is the most accurate system, however it is interesting to see that while the rightmost column is least accurate, the model correctly directs its attention to the most informative areas (dog faces) but lacks enough information to classify correctly. It is also interesting to note in the last sample, the rightmost model correctly classifies a breed given a non-face feature, showing that the system has learned to identify a variety of useful parts instead of relying solely on facial features.

\end{document}